\def\year{2018}\relax
\documentclass[letterpaper]{article}
\usepackage{aaai18}
\usepackage{times}
\usepackage{helvet}
\usepackage{courier}
\usepackage{amsfonts}
\usepackage{amsmath,amssymb,amscd,epsfig,amsfonts,rotating}
\usepackage{graphicx}
\usepackage{epstopdf}
\usepackage{subfigure}
\usepackage{multirow}
\usepackage{booktabs}
\usepackage{xcolor}
\usepackage{url}
\usepackage{amsmath,amscd,amsbsy,amssymb,latexsym,url,bm}
\usepackage{epsfig,epstopdf,graphicx,subfigure}
\usepackage{enumitem,balance}
\usepackage{wrapfig}
\usepackage{mathrsfs, euscript}
\usepackage{algorithm}
\usepackage{algorithmic}
\usepackage{amssymb}
\usepackage{ifpdf}
\usepackage{makecell}
\usepackage{tikz}
\usepackage{appendix}
\usetikzlibrary{matrix,chains,positioning,decorations.pathreplacing,arrows}

\newcommand{\vect}[1]{\mathbf{#1}}

\begin{document}
%
\title{DeepTransport: Learning Spatial-Temporal Dependency \\
	for Traffic Condition Forecasting}
\author{
		Xingyi Cheng\thanks{This work was done before leaving Baidu. Email: derrickzy@gmail.com.}, Ruiqing Zhang, Jie Zhou, Wei Xu\\
		Baidu Research - Institue of Deep Learning \\
		derrickzy@gmail.com\\
		\{zhangruiqing01,zhoujie01,wei.xu\}@baidu.com		
}
\maketitle
\begin{abstract}
Predicting traffic conditions has been recently explored as a way to relieve traffic congestion. Several pioneering approaches have been proposed based on traffic observations of the target location as well as its adjacent regions, but they obtain somewhat limited accuracy due to a lack of mining road topology. To address the effect attenuation problem, we suggest taking into account the traffic of surrounding locations(wider than the adjacent range). We propose an end-to-end framework called DeepTransport, in which Convolutional Neural Networks (CNN) and Recurrent Neural Networks (RNN) are utilized to obtain spatial-temporal traffic information within a transport network topology. In addition, an attention mechanism is introduced to align spatial and temporal information. Moreover, we constructed and released a real-world large traffic condition dataset with a 5-minute resolution.  Our experiments on this dataset demonstrate our method captures the complex relationship in the temporal and spatial domains. It significantly outperforms traditional statistical methods and a state-of-the-art deep learning method. 

\end{abstract}
\section{Introduction}
\begin{figure}
\centering
\includegraphics[width=60mm]{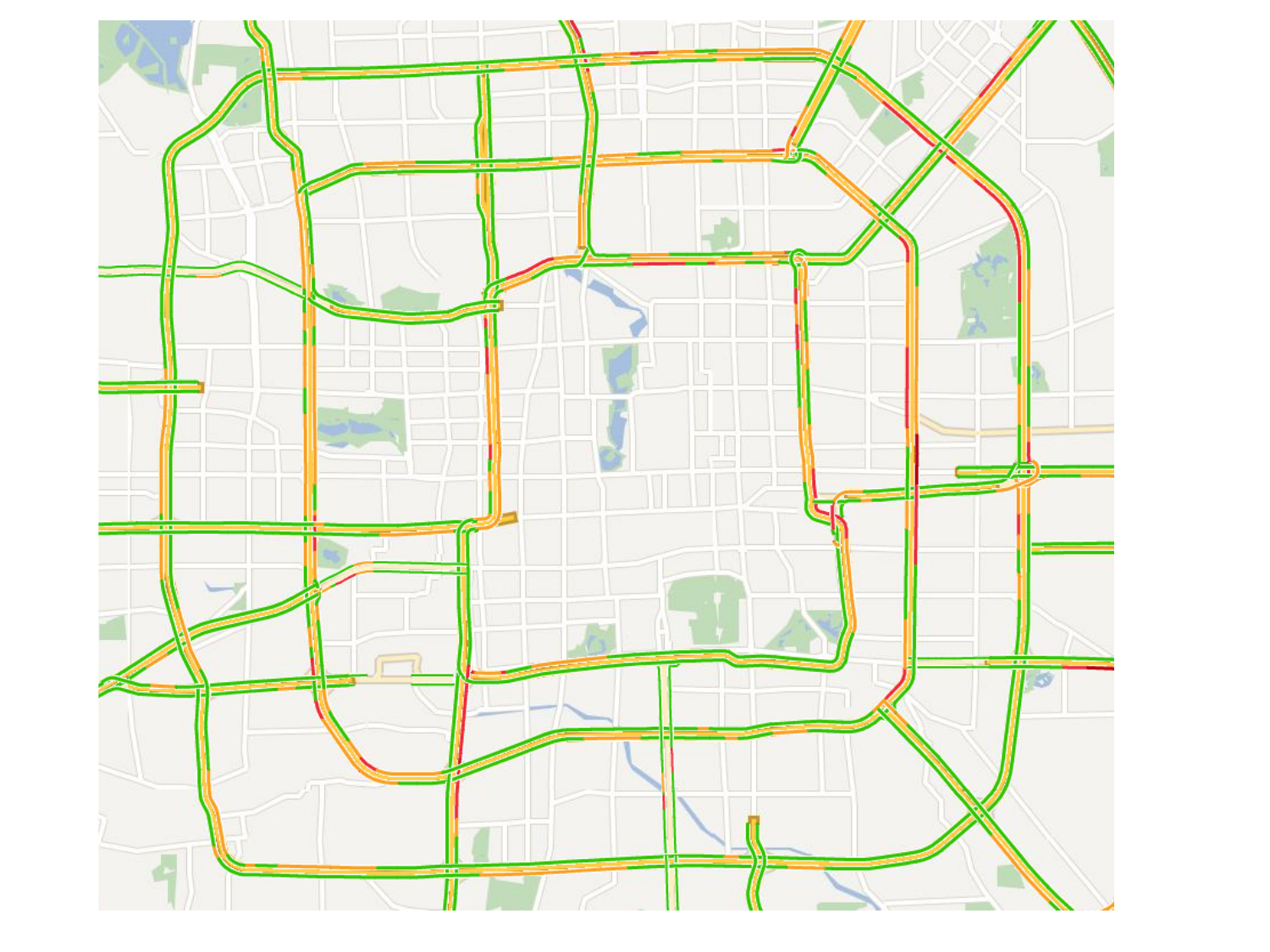}
\caption{A real-time traffic network example from a commercial map app, the networks including many locations and the color~(green, yellow, red, dark red) depth illustrated the condition of a location(a stretch of road).
}
\label{fig::map}
\end{figure}

With the development of location acquisition and wireless devices, a vast amount of data with spatial transport networks and timestamps can be collected by mobile phone map app. The majority of map apps can tell users real-time traffic conditions, as shown in Figure~\ref{fig::map}. However, only the current traffic conditions are not enough for making effective route planning, a traffic system to predict future road conditions may be more valuable.

In the past, there are mainly two approaches for traffic prediction: time-series analysis based on classical statistics and data-driven methods based on machine learning. Most former methods are univariate; they predict the traffic of a place at a certain time. The fundamental work was Auto Regressive Integrated Moving Average~(ARIMA)~\cite{ahmed1979analysis} and its variations~\cite{BeiPan:harima,williams:sarima}. Motivated by the fact~\cite{williams:arimax} that traffic evolution is a temporal-spatial phenomenon, multivariate methods with both temporal and spatial features were proposed. ~\cite{stathopoulos2003multivariate} developed a model that feeds on data from upstream detectors to improve the predictions of downstream locations.  However, many statistics are needed in such methods. On the other hand, data-driven methods~\cite{jeong2013supervised,vlahogianni2005optimized} fit a single model from vector-valued observations including historical scalar measurements with the trend, seasonal, cyclical, and calendar variations. For instance, ~\cite{deng2016latent} expressed traffic patterns by mapping road attributes to a latent space. However, the linear model here is limited in its ability to extract effective features.

Neural networks and deep learning have been demonstrated as a unified learning framework for feature extraction and data modeling. Since its applicability in this topic, significant progress has been made in related work. Firstly, both temporal and spatial dependencies between observations in time and space are complex and can be strongly nonlinear. While the statistics frequently fail when dealing with nonlinearity, neural networks are powerful to capture very complex relations~\cite{lecun2015deep}. Secondly, neural networks can be trained with raw data in an end-to-end manner. Apparently, hand-crafted engineered features that extract all information from data spread in time and space are laborious. Data-driven based neural networks extract features without the need for statistical features. e.g., Mean or variance of all adjacent locations of the current location.  The advantage of neural networks for traffic prediction has long been discovered by researchers. Some early work~\cite{chang1995predicting,innamaa2000short} simply put observations into the input layer, or take sequential features into consideration ~\cite{dia2001object} to capture temporal patterns in time-series. Until the last few years, some works of deep learning were applied. For instance, Deep Belief Networks (DBN)~\cite{huang2014deep} and Stack Autoencoders~(SAEs)~\cite{lv2015traffic}. However, input data in these works are directly concatenated from different locations, which ignored the spatial relationship. 
In general, the existing methods are either concerned with the time series or just a little use of the spatial information. Depending on traffic conditions of a ``narrow'' spatial range will undoubtedly degrade prediction accuracy. To achieve a better understanding of spatial information, we propose to solve this problem by taking the intricate topological graph as a key feature in traffic condition forecasting, especially for long prediction horizons.

To any target location as the center of radiation, surrounding locations with the same order form a ``width'' region, and regions with different order constitute a ``depth'' sequence. We propose a double sequential deep learning model to explore the traffic condition pattern. This model adopts a combination of convolutional neural networks (CNN)~\cite{lecun1995convolutional} and recurrent networks with long short-term memory (LSTM) units~\cite{hochreiter1997long} to deal with spatial dependencies. CNN is responsible for maintaining the ``width'' structure, while LSTM for the ``depth'' structure. To depict the complicated spatial dependency, we utilize the attention mechanism to demonstrate the relationships between time and space.

The main contribution of the paper is summarized as follows:
\begin{itemize}
	\item We introduce a novel deep architecture to enable temporal and dynamical spatial modeling for traffic condition forecasting. 
\end{itemize}

\begin{itemize}
	\item We propose the necessity of aligning spatial and temporal information and introduce attention mechanism into the model to quantify their relationship. The obtained attention weight is helpful for daily traveling and path planning.
\end{itemize}

\begin{itemize}
	\item  Experiment results demonstrate that the proposed model significantly outperforms existing methods based on deep learning and time series forecasting methods.

\end{itemize}

\begin{itemize}
	\item We also release a real large (millions) traffic dataset with topological networks and temporal traffic conditions~\footnote{{https://github.com/cxysteven/MapBJ}} for ASC Student Supercomputer Challenge 2017~(ASC17), which was developed on PaddlePaddle platform~\footnote{https://github.com/PaddlePaddle/Paddle}.
\end{itemize}

\section{Preliminary}
In this section, we briefly revisit the traffic prediction problem and introduce notations in this work.
\subsection{Common Notations and Definition}

\begin{figure}
\begin{minipage}[b]{.5\linewidth}
\centering
\subfigure[A plain graph at a time point]{
\includegraphics[width=4cm]{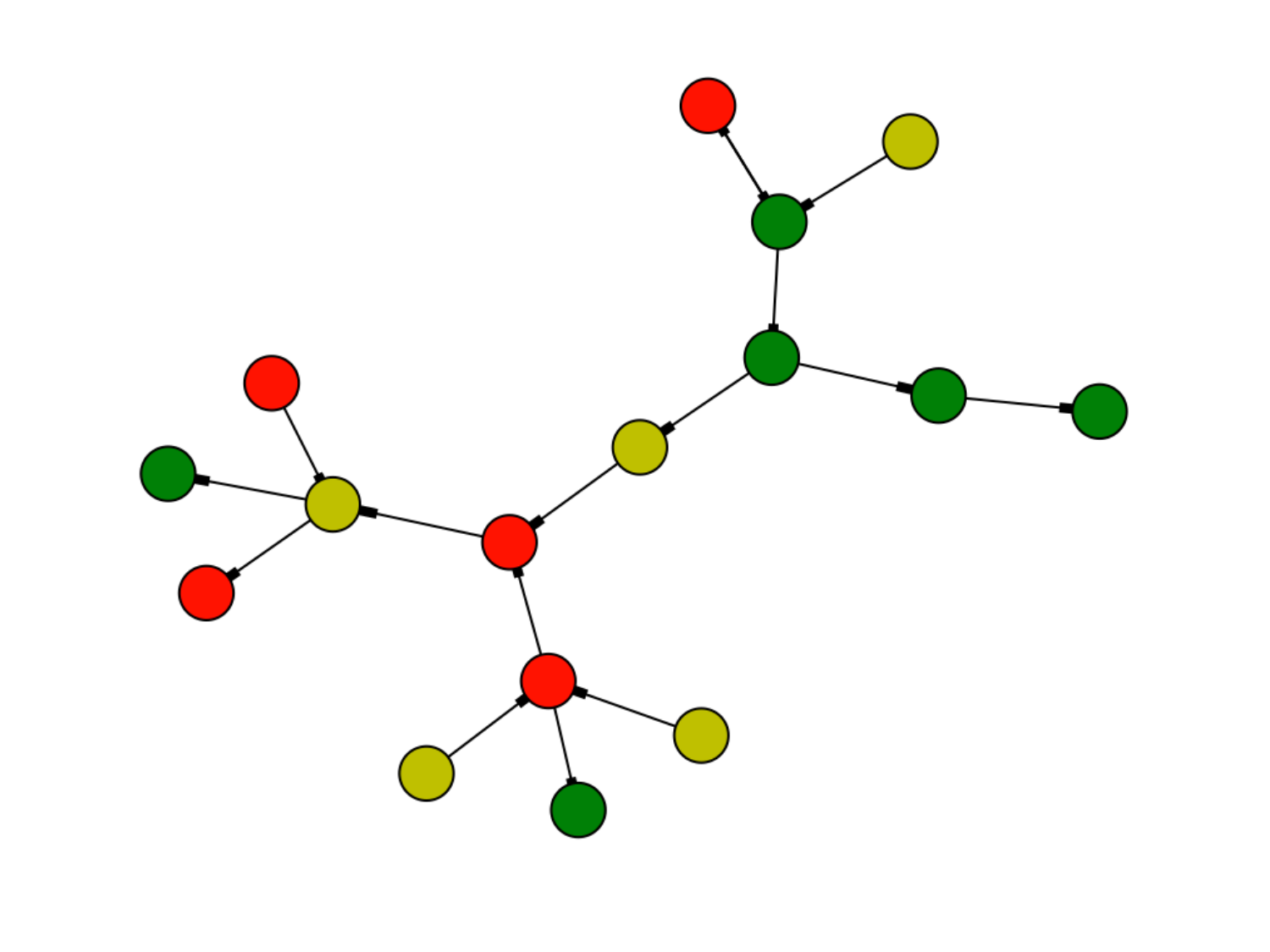}
\label{fig::graph1}
}
\end{minipage}%
\begin{minipage}[b]{.5\linewidth}
\centering
\subfigure[A graph with time-series] {
\includegraphics[width=4cm]{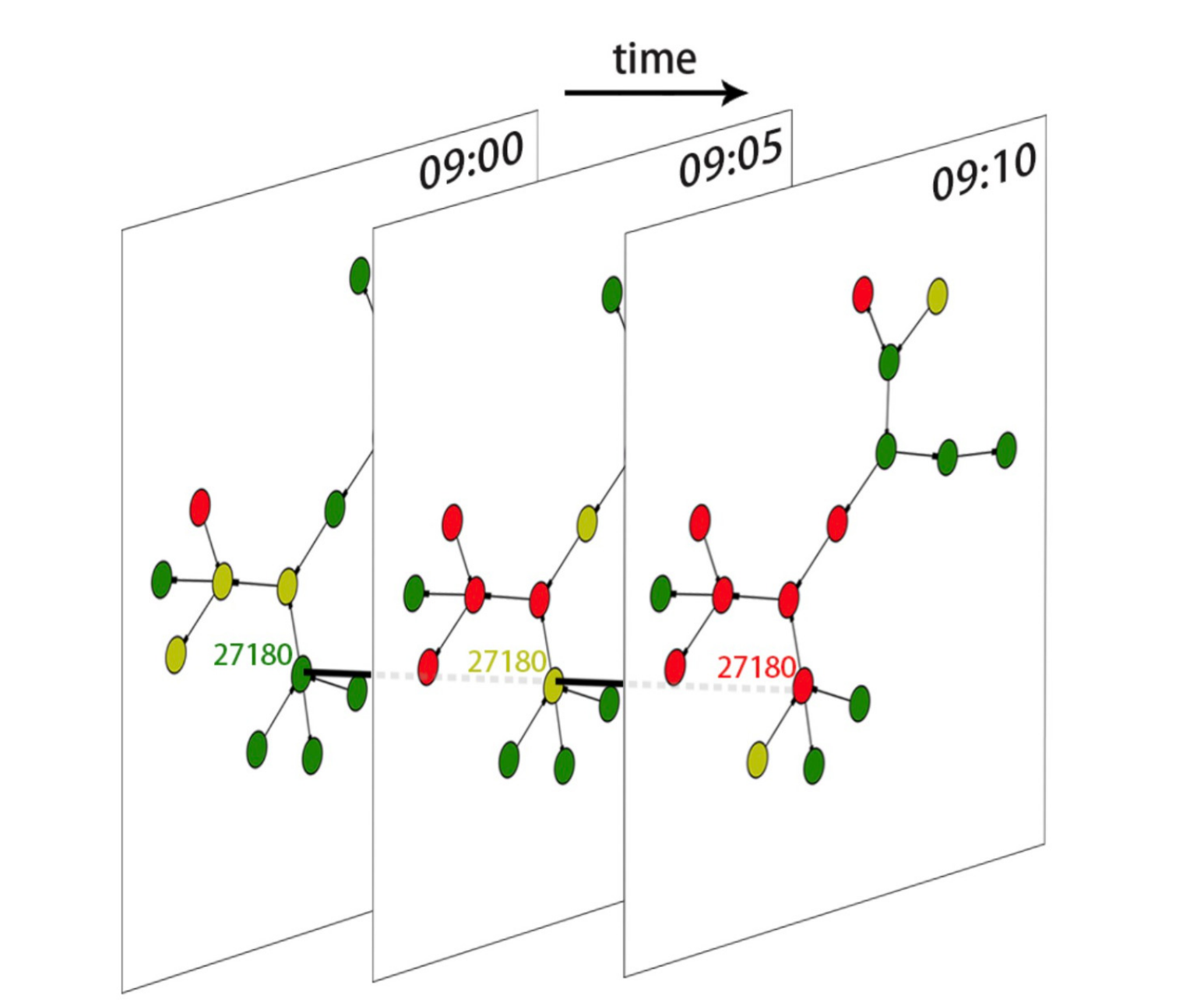}
\label{fig::graph2}
}
\end{minipage}
\caption{traffic condition. Five colors in this graph denote five states for visually displaying: green(1, fluency), yellow(2, slow), red(3, congestion), and dark red(4, extreme congestion). ``27180'' is the ID number of a location(road section).}
\label{fig::simple_graph}
\end{figure}

A traffic network can be represented in a graph in two ways. Either monitoring the traffic flow of crossings, taking the crossing as a node and road as an edge of the graph, or conversely, monitoring the condition of roads, take roads as nodes and crossings as connecting edges. The latter annotation is adopted in our work. Taking figure~\ref{fig::graph1} as an example, each colored node corresponds to a stretch of road in a map app.

We consider a graph consisting of weighted vertices and directed edges. Denote the graph as $G = \langle V, E \rangle$. $V$ is the set of vertices and $E \subseteq \{(u, v) \vert u \in V, v \in V\}$ is the set of edges, where (u, v) is an ordered pair. A location(vertex) $v$ at any time point $t$ has five traffic condition states $c(v, t) \in \{0, 1, 2, 3, 4\} $, expressing not-released, fluency, slow, congestion, extreme congestion respectively. Figure~\ref{fig::graph2} presents an example of road traffic at three-time points in an area.

\textbf{Observations}:
Each vertex in the graph is associated with a feature vector, which consists of two parts, \textit{time-varying} $O$ and \textit{time-invariant} variables $F$. Time-varying variables that characterize the traffic network dynamically are traffic flow observations aggregated by a 5-minute interval. Time-invariant variables are static features as natural properties which do not change with time s, such as the number of input and output degrees of a road, its length, limit speed, and so forth.

In particular, the time-varying and time-invariant variables are denoted as:

\begin{equation}
\centering {
\begin{matrix}
\vect{O}_{v, t} = \begin{bmatrix}
c(v, t) \\
c(v, t-1) \\
\vdots   \\
c(v, t-p) 
\end{bmatrix} &
\vect{F}_{v} = \begin{bmatrix} 
f_{v, 1} \\
f_{v, 2} \\
\vdots \\
f_{v, k}
\end{bmatrix} &
\end{matrix}
}
\end{equation} 

where $c(v, t)$ is traffic condition of vertex $v$ at time $t$, $p$ is  the length of historical measurement. $ f_{v,k}$ are time-invariant features.

\begin{figure}
\begin{minipage}{0.5\linewidth}
\centering
\subfigure[The direct graph of a location.]{
\includegraphics[width=3.5cm]{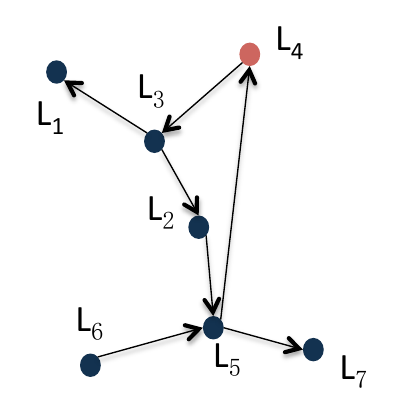}
\label{fig::simple_graph}
}
\end{minipage}%
\begin{minipage}{0.5\linewidth}
\centering
\subfigure[Upstream flow and downstream flow neighbor of $L_4$ with in order 2.]{
\includegraphics[width=4cm]{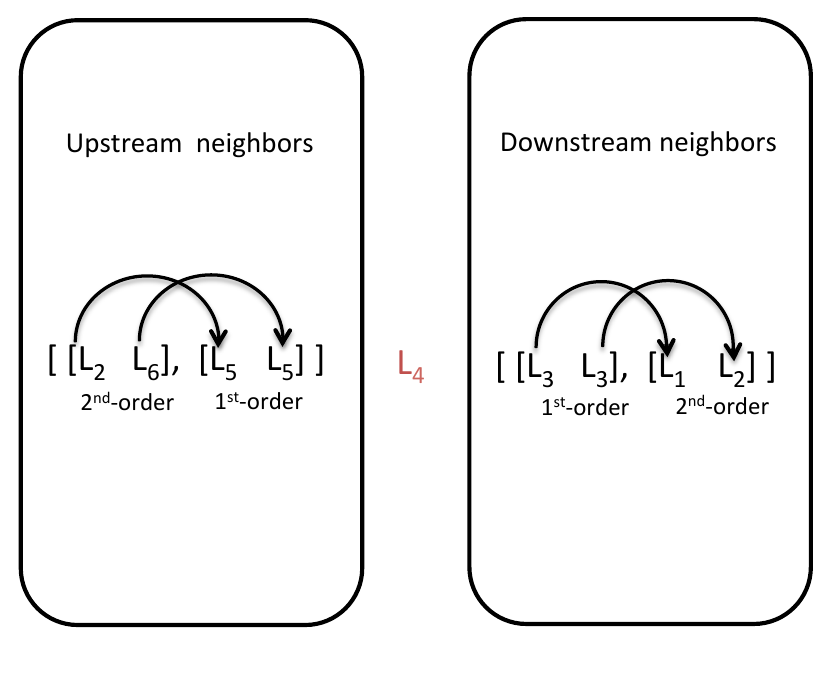}
\label{fig::input}
}
\end{minipage}
\caption{An example of a directed graph and order slot notation in DeepTransport}
\label{fig::exp}
\end{figure}

\textbf{Order Slot}:
In a path of the directed graph, the number of edges required to take from one vertex to another is called order. Vertices of the same order constitute an order slot. Directly linked vertices are termed first-order neighbors. Second-order spatial neighbors of a vertex are the first-order neighbors of its first-order neighbors and so forth. For any vertex in our directed graph, we define the incoming traffic flow as its upstream flow and the outflow as its downstream flow. Take figure~\ref{fig::simple_graph} as an example, $L_4$ is the target location to be predict. $L_3$ is the first-order downstream vertex of $L_4$. $L_1, L_2$ is the first order downstream set of $L_3$ and they constitute the second order slot of $L_4$. Each vertex in the traffic flow that goes in one direction is affected by its upstream flow and downstream flow. The first and second order slots of $L_4$ is shown in Figure~\ref{fig::input}. 
Introducing the dimension of time series, any location $L_{v,t}$ is composed of two vectors, $O_{v,t}$ and $F_{v}$. Any order slot consists of some locations:

\begin{equation}
\centering {
\begin{matrix}
\vect{L}_{v, t} = \begin{bmatrix}
\vect{O}_{v, t} \\ 
\vect{F}_{v} 
\end{bmatrix} &
\vect{X}^j_{v, t} = \begin{bmatrix}
\vect{L}^T_{u_1, t} \\
\vect{L}^T_{u_2, t} \\
\vdots   \\
\vect{L}^T_{u_k, t} 
\end{bmatrix}
\end{matrix}
}
\end{equation} 

where location index $u_{\cdot}$ is one of the $j$th order neighbors of $v$.

\textbf{Perceptive Radius}:
The maximum ordered number controls the perceptive scope of the target location. It is an important hyperparameter describing spatial information, we call it perceptive radius and denote it as $r$. 

\textbf{Problem Definition}:
According to the above notation, we define the problem as follows: Predict a sequence of traffic flow $\vect{L}_{v, t+h}$ for prediction horizon $h$ given the historical observations of ${\vect{L}}_{v', t'}$, where $v' \in neighbor(v, r)$, $t' \in \{t-p,\cdots, t\}$, $r \in \{0,\cdots,R\}$ is perceptive radius and $p$ is the length of historical measurement.

\section{Model}
\begin{figure*}[!ht]
\centering
\includegraphics[width=7in, height=5.25in]{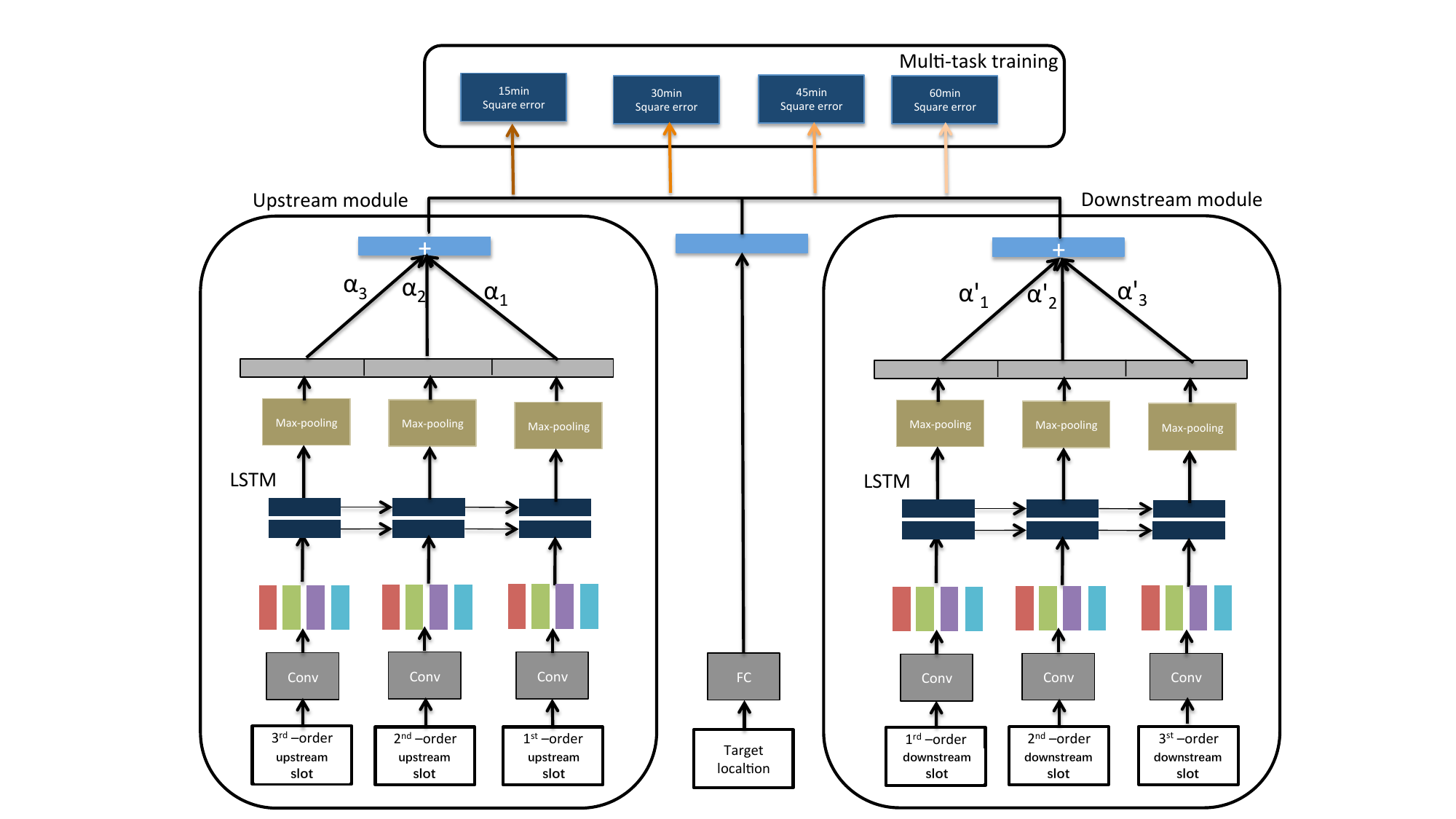}
\caption{An example of the model architecture. There are three slots in upstream and downstream module respectively, each with input vertices length of two. The convolution operation has four sharing feature map. The middle block demonstrates that the target location is propagated by the fully-connected operation. A multi-task module that with four cost layers on the top block. Conv: Convolution; FC: Fully-connection.}
\label{fig::DeepTransport}
\end{figure*}

As shown in Figure~\ref{fig::DeepTransport}, our model consists of four parts: upstream flow observation(left), target location module(middle), downstream flow observation(right), and training cost module(top). In this section, we detail the work process of each module.

\subsection{Spatial-temporal Relation Construction}

Since the traffic condition of a road is strongly impacted by its upstream and downstream flow,  we use a convolutional subnetwork and a recurrent subnetwork to maintain the road topology in the proposed model.

\subsubsection{Convolutional Layer}


CNN is used to extract temporal and ``width'' spatial information. As demonstrated in the example of figure~\ref{fig::exp}, when feeding into our model, $L_4$'s first upstream neighbor $L_5$ should be copied twice, because there are two paths to $L_4$,  that are $[L_6, L_5]$ and $[L_2, L_5]$. With the exponential growth of paths, the model suffers from high dimension and intensive computation. Therefore, we employ a convolution operation with multiple encoders and shared weights~\cite{lecun1995convolutional}. To further reduce the parameter space while maintaining independence among vertices with the same order, we set the convolution stride to the convolution kernel window size, which is equal to the length of a vertex's observation representation.

The non-linear convolutional feature is obtained as follows:
\begin{eqnarray}
\vect{e}^{r}_{up, q}&=&\sigma(\vect{W}_{up, q} * {\vect{U}_{v, t}} + \vect{b}_{up, q}), \\
\vect{e}^{r}_{down, q}&=&\sigma(\vect{W}_{down, q} * {\vect{D}_{v, t}} + \vect{b}_{down, q}), 
\end{eqnarray}
where $\vect{U}_{v, t} = [\vect{X}^1_{v,t},\cdots,\vect{X}^r_{v,t}]$(only upstream neighbors) is denoted as upstream input matrix, while $\vect{D}_{v, t}$ is downstream input matrix. The $\vect{e}^{r}_{\cdot, q}$ is at $r$th order vector of upstream or downstream module where $q \in \{1, 2...m\} $ and $m$ is the number of feature map. We set $\vect{e}^{r}_{up} = [ \vect{e}^{r}_{up, 1}, \cdots, \vect{e}^{r}_{up, m} ] $ and $\vect{e}^{r}_{up} \in \mathbb{R}^{l \times m} $,  $l$ is the number of observations in a slot. Similarly, we can get the $\vect{e}^{r}_{down}$. The weights $\vect{W} $ and bias $\vect{b}$ composes parameters of CNN subnetworks. $\sigma$ represents nonlinear activation, we empirically adopt the $\text{tanh}$ function here.

\subsubsection{Recurrent Layer}
RNN is utilized to represent each path that goes to the target location(upstream path) or goes out from the target location(downstream path). The use of RNN has been investigated for traffic prediction for a long time, ~\cite{Dia2001An} used a Time-Lag RNN for short-term speed prediction(from 20 seconds to 15 minutes), and ~\cite{Lint2002Freeway} adopted RNN to model state space dynamics for travel time prediction. In our proposed method, since the upstream flow is from high-order to low-order, while the downstream flow is contrary, the output of the CNN layer in the upstream module and downstream module is fed into RNN layer separately. 
 
The structure of vehicle flow direction uses LSTM with ``peephole'' connections to encode a path as a sequential representation. In LSTM, the forget gate $\vect{f}$ controls memory cell $\vect{c}$ to erase, the input gate $\vect{i}$ helps to ingest new information, and the output gate $\vect{o}$ exposes the internal memory state outward. Specifically, given a $r$th slot matrix $\vect{e}^r_{down} \in \mathbb{R}^{l \times m}$, map it to a hidden representation $\vect{h}^r_{down} \in \mathbb{R}^{l \times d}$ with LSTM as follows:
\begin{align}
	\begin{bmatrix}
		\mathbf{\tilde{c}}^{r} \\
		\mathbf{o}^{r} \\
		\mathbf{i}^{r} \\
		\mathbf{f}^{r}
	\end{bmatrix}
	&=
	\begin{bmatrix}
		\tanh \\
		\sigma \\
		\sigma \\
		\sigma
	\end{bmatrix}
    \begin{pmatrix}
	\mathbf{W}_p
	\begin{bmatrix}
		\mathbf{e}^{r} \\
		\mathbf{h}^{r-1} \\
	\end{bmatrix}+\mathbf{b}_p
    \end{pmatrix}, \label{eq:lstm1}\\
    \mathbf{c}^{r} &=  \mathbf{\tilde{c}}^{r} \odot \mathbf{i}^{r} + \mathbf{c}^{r-1} \odot \mathbf{f}^{r}, \label{eq:lstm2} \\
         \mathbf{h}^{r} &= [\mathbf{o}^{r}  \odot \tanh\left( \mathbf{c}^{r}  \right)]^T\label{eq:lstm3},
\end{align}
where $\mathbf{e}^r \in \mathbb{R}^{l\times m}$ is the input at the $r$th order step; $\mathbf{W}_p \in \mathbb{R}^{4d\times (m+d)}$ and $\mathbf{b}_p \in \mathbb{R}^{4d}$ are parameters of affine transformation;
$\sigma$ denotes the logistic sigmoid function and $\odot$ denotes elementwise multiplication.

The update of upstream and downstream LSTM units can be written precisely as follows:
\begin{align}
\mathbf{h}^r_{down} &= \mathbf{LSTM}(\mathbf{h}^{r-1}_{down},\mathbf{e}^r_{down}, \theta_p).  \label{eq:LSTM}
\end{align}
\begin{align}
\mathbf{h}^r_{up} &= \mathbf{LSTM}(\mathbf{h}^{r+1}_{up},\mathbf{e}^r_{up}, \theta_p).  \label{eq:LSTM}
\end{align}

The function $\mathbf{LSTM}(\cdot, \cdot, \cdot)$ is a shorthand for Eq. (\ref{eq:lstm1}-\ref{eq:lstm3}), in which $\theta_p$ represents all the parameters of $\mathbf{LSTM}$.

\subsubsection{Slot Attention}
To get the representation of each order slot, max-pooling is performed on the output of LSTM. As $\mathbf{h}^r$ represents the status sequence of the vertices in the corresponding order slot, we pool on each order slot to get $r$ number of slot embeddings $\vect{S}_{up} = [ \vect{s}^{1}_{up}, \cdots, \vect{s}^{r}_{up} ]$ and $\vect{S}_{down} = [ \vect{s}^{1}_{down}, \cdots, \vect{s}^{r}_{down} ]$. Since different order slots have different effects on target prediction, we introduce attention mechanisms to align these embeddings. Given the target location hidden representation $\vect{g}$, we get the $j$th slot attention weights~\cite{Bahdanau2014Neural,Rockt2015Reasoning} as follows:
\begin{align}
	\alpha_j & = \frac{\exp{a(\mathbf{g},\mathbf{s}^{j})}}{\sum_{k=1}^{r}\exp{a(\mathbf{g},\mathbf{s}^{k})}}. \label{eq:attention} 
\end{align}

We parametrize the model $a$ as a Feedforward Neural Network that is used to compute the relevance between the target location and the corresponding order slot. The weight $\alpha_j$ is normalized by a softmax function. To write it precisely, we let  $\mathbf{ATTW}(\mathbf{s}^j)$ as a shorthand for Eq.(\ref{eq:attention}), we get the upstream and downstream hidden representation by weighting the sum of these slots:
\begin{align}
\mathbf{z}_{down} = \sum_{j=1}^{r} \mathbf{ATTW}(\mathbf{s}^j_{down}) \mathbf{s}^j_{down}. 
\end{align}
\begin{align}
\mathbf{z}_{up} = \sum_{j=1}^{r} \mathbf{ATTW}(\mathbf{s}^j_{up}) \mathbf{s}^j_{up}. \end{align}

Lastly, we concatenate the $\mathbf{z}_{up}$, $\mathbf{z}_{down}$ and the target location's hidden representation $\mathbf{g}$ and then sent them to the cost layer. 




\subsection{Top Layers with Multi-task Learning}
The choice of cost function on the top layer is tightly coupled with the choice of the output unit. We simply use square error to fit the future conditions of the target locations.

Multi-task learning is first introduced by ~\cite{huang2014deep} for traffic forecasting tasks. It is considered as soft constraint imposed on the parameters arising out of several tasks~\cite{evgeniou2004regularized}. These  Additional training examples put more pressure on the parameters of the model towards values that generalize well when part of a model is shared across tasks. Forecasting traffic future conditions is a multi-task problem as time goes on and different time points correspond to different tasks.  In the DeepTransport model, in addition to the computation of the attention weights and affine transformations of the output layer, all other parameters are shared.

\section{Experiments} 
\subsection{Dataset}
We adopt \textit{snowball sampling method}~\cite{biernacki1981snowball} to collect an urban areal dataset in Beijing from a commercial map app and named it ``MapBJ''. The dataset provides traffic conditions in $\{$fluency, slow, congestion, extreme congestion$\}$. The dataset contains about 349 locations which are collected from March 2016 to June every five minutes. We select the first two months' data for training and the remaining half month for testing. Besides traffic topological graphs and time-varying traffic conditions, we also provide the limit speed of each road. Since the limit speed of different roads may be very distinct, and location segmentations method regards this as an important reference index. We introduce a time-invariable feature called \text{limit level} and discretize it into four classes.

\subsection{Evaluation}
Evaluation is ranked based on \textit{quadratic weighted Cohen's Kappa}~\cite{ben2008comparison}, a criterion for evaluating the performance of categorical sorting. 

In our problem, \textit{quadratic weighted Cohen's Kappa} is characterized by three $4 \times 4$ matrices: observed matrix $\vect{O}$, expected matrix $\vect{E}$ and weight matrix $\vect{w}$. Given Rater $\vect{A}$(ground truth) and Rater $\vect{B}$(prediction), $\vect{O}_{i,j}$  denotes the number of records rating $i$ in $\vect{A}$ while rating $j$ in $\vect{B}$, $\vect{E}_{i,j}$ indicates how many samples with label $i$ is expected to be rated as $j$ by $\vect{B}$ and $\vect{w}_{i,j}$ is the weight of different rating, 
\begin{eqnarray}
\vect{w}_{i,j} = \frac{(i-j)^2}{(N-1)^2},
\end{eqnarray}
where N is the number of subjects, we have $N=4$ in our problem. From these three matrices, the quadratic weighted kappa is calculated as:
\begin{eqnarray}
\kappa = 1 - \frac{\Sigma_{i,j}{\vect{w}_{i,j}\vect{O}_{i,j}}}{\Sigma_{i,j}{\vect{w}_{i,j}\vect{E}_{i,j}}}.
\end{eqnarray}
This metric is typically in the range of 0 (random agreement between raters) to 1 (complete agreement between raters).

\subsection{Implementation Details}
We use the open-source deep learning platform \textit{PaddlePaddle} for the implementation and experiments. PaddlePaddle has two important files for running the program: \textit{data providers} and \textit{trainer configuration}. \textit{data providers} is usually used for data preprocessing with Python language and \textit{trainer configuration} is responsible for parameter setting and building neural networks layer by layer.
Since the condition value ranges in $\{1, 2, 3, 4\}$, the multi-classification loss can be treated as the objective function. However, the cost layer with softmax cross-entropy does not take into account the magnitude of the rating. Thus, square error loss is applied as the training objective. But another disadvantage of the straightforward use of linear regression is that the predicted value may be out of the range in $\{1, 2, 3, 4\}$. However, we can avoid this problem by labeling projection as follows:

We have a statistical analysis on the state distribution of training data. Fluency occupies 88.2\% of all records, fluency and slower occupies about 96.7\%,  fluency, slower and congestion occupies about 99.5\%, the extreme congestion is very rare that it accounts for only 0.5\%. Therefore, we rank the prediction result in ascending order and set the first 88.2\% to fluency, 88.2\%-96.7\% to slower, 96.7\%-99.5\% to congestion, and 99.5\%-100\% to extreme congestion.

We put all the observations into $32$ dimension continuous vectors. The training optimization is optimized by back-propagation using Adam~\cite{kingma2014adam}. Parameters are initialized with uniformly distributed random variables and we use batch size 1100 for 11 CPU threads, with each thread processing 100 records. All models are trained until convergence. Besides, there are two important hyperparameters in our model,  the length of historical measurement $p$ and perceptive radius $r$ that control temporal and spatial magnitude respectively.

\subsection{Choosing Hyperparamerters} 
We intuitively suppose that expanding perceptive radius would improve prediction accuracy, but also increase the amount of computation, so it is necessary to explore the correlation between the target location and its corresponding $r$th order neighbors.

Mutual Infomation(MI) measures the degree of correlation between two random variables. When MI is 0, it means the given two random variables are completely irrelevant. When MI reaches the maximum value, it equals to the entropy of one of them, and the uncertainty of the other variable can be eliminated. MI is defined as
\begin{eqnarray}
\centering
MI(\vect{X};\vect{Y}) &=& H(\vect{X}) - H(\vect{X} \vert \vect{Y}) \nonumber \\
       &=& \sum_{x \in \vect{X}, y \in \vect{Y}}p(x,y)\log\frac{p(x,y)}{p(x)p(y)},
\end{eqnarray}
where $H(\vect{X})$ and $H(\vect{X}|\vect{Y}) $ are marginal entropy and conditional entropy respectively. MI describes how much uncertainty is reduced.

With MI divided by the average of entropy of the given two variables, we get Normalized mutual information(NMI) in $[0,1]$:
\begin{eqnarray}
\centering
NMI(\vect{X};\vect{Y}) &=& 2\frac{MI(\vect{X},\vect{Y})}{H(\vect{X}) + H(\vect{Y})}.
\end{eqnarray}
We calculated NMI between the observation of each vertex and its $r$th neighbors over all time points. The NMI gradually decreases as the order increases, it values  0.116, 0.052, 0.038, 0.035, 0.034 for $r$ in $\{1,2,3,4,5\}$ respectively and hardly change after $r > 5$.


Therefore, we set the two hyperparameters as $p \in \{3, 6, 12, 18\}$ (corresponding to 15, 30, 60, 90 minutes past measurements as 5-minute record interval) and $r \in \{1, 2, 3, 4, 5\}$.

\subsection{Effects of Hyperparameters}



\begin{figure}
\begin{minipage}[b]{.5\linewidth}
\centering
\subfigure[Prediction with $p=12$]{
\includegraphics[width=4cm]{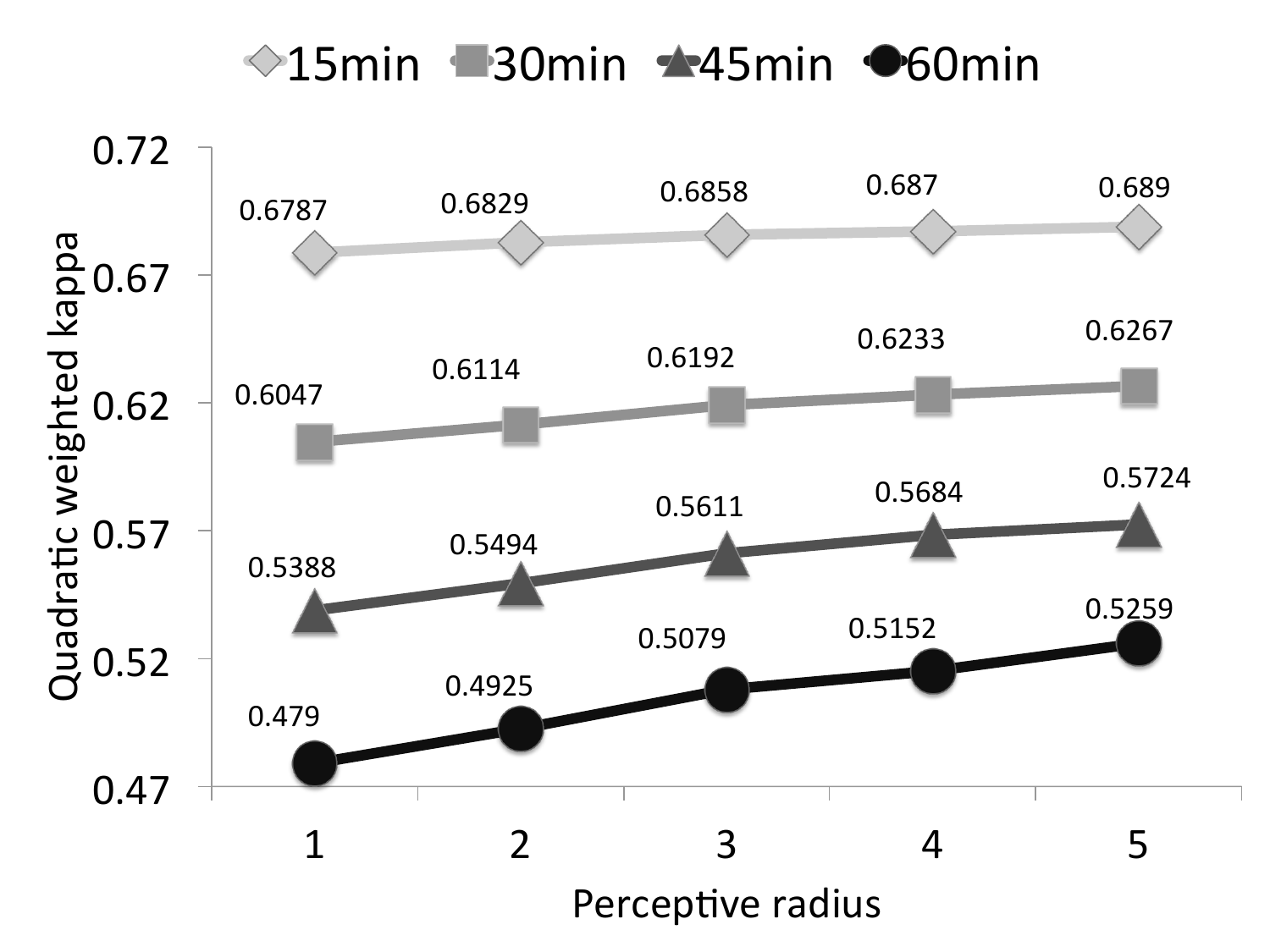}
\label{fig::h15}
}
\end{minipage}%
\begin{minipage}[b]{.5\linewidth}
\centering
\subfigure[60-minute prediction] {
\includegraphics[width=4cm]{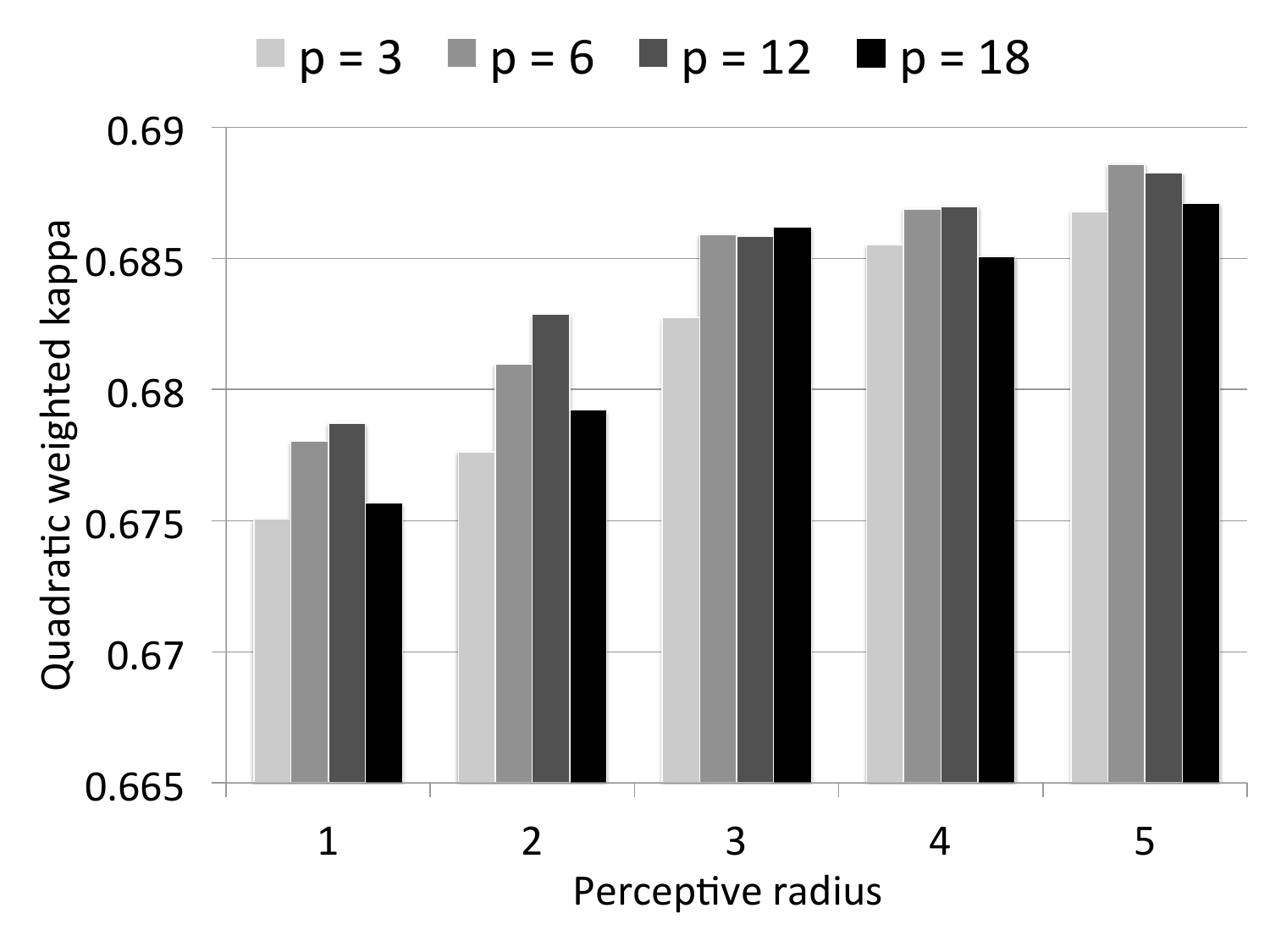}
\label{fig::h45}
}
\end{minipage}
\caption{Averaged quadratic weighted kappa over the perceptive radius $r$ and the length of historical measurement $p$ on validation data. The left figure illustrates that as a function of perceptive radius $r$ increases, the longer horizon prediction has more growth. The right figure shows that the optimal $p$ should be chosen by observing the perceptive radius.}
\label{fig::worse}
\end{figure}


Figure~\ref{fig::worse} shows the averaged quadratic weighted kappa of the corresponding prediction horizon. Figure~\ref{fig::h15} illustrates 1) a closer prediction horizon always performs better; 2) As $r$ increases, its impaction on the prediction also increases. This can be seen from the slope between $r=1$ and $r=5$, the slope at 60-min is greater than the same segment of 15-min. Figure~\ref{fig::h45} takes a 60-min estimation as an example, indicating that the predictive effect is not monotonically increasing as the length of measurement $p$, and the same result can be obtained at other time points. This is because the increase in $p$ brings an increase in the number of parameters, which leads to overfitting.

\subsection{Comparison with Other Methods}
We compared  DeepTransport with four representative approaches: Random Walk(RW), Autoregressive Integrated Moving Average(ARIMA) and Stacked AutoEncoders(SAEs).


\textbf{RW}:
In this baseline, the traffic condition at the next moment is estimated as a result of the random walk at the current moment condition that adds a white noise(a normal variable with zero mean and variance one). 

\textbf{ARIMA}:
It~\cite{ahmed1979analysis} is a common statistical method for learning and predicting future values with time series data. We take a grid search over all admissible values of $p$, $d$ and $q$ which are less than $p$ = 5, $d$ = 2 and $q$ = 5.

\textbf{FNN}:
We also implemented Feed-forward Neural Networks (FNN), with a single hidden layer and an output layer with regression cost. The hidden layer has 32 neurons, and four output neurons refer to the prediction horizon. Hyperbolic tangent function and linear transfer function are used for the activation function and output respectively.

\textbf{SAEs}:
We also implemented SAEs~\cite{lv2015traffic}, one of the most effective deep learning-based methods for traffic condition forecasting. It concatenates observations of all locations to a large vector as inputs. SAEs also can be viewed as a pre-training version of FNN with a large input vector proposed by~\cite{Polson2017Deep}. The stacked autoencoder is configured with four layers with [256, 256, 256, 256] hidden units for pre-train. After that, a multi-task linear regression model is trained on the top layer.

 
Besides, we also provide the result of DeepTransport with two configurations, with $r=1, p=12$ (DeepTransport-R1P12) and $r=5, p=12$ (DeepTransport-R5P12).

Table~\ref{tab::comp} shows the results of our model and other baselines on MapBJ. In summary, the models that use spatial information(SAEs, DeepTransport) significantly have higher performance than those that do not use(RW, ARIMA, FNN), especially in longer prediction horizons. On the other hand, SAEs is a fully-connected form, meaning that it assumes that any couple locations directly connect to each other so it neglects the topology structure of transport networks. On the contrary, DeepTransport considers traffic structure results as higher performance than these baselines, demonstrating that our proposed model has good generalization performance.

\begin{table}[]
	\centering

	\resizebox{1\linewidth}{!}{
		\begin{tabular}{c c c c c c} \hline
			\, & \multicolumn{4}{c}{Quadratic Weighted Kappa} \\ \hline
			Model & 15-min  & 30-min & 45-min & 60-min & Avg.\\ \hline
			RW  & 0.5106&0.4474 &0.3917& 0.3427  & 0.4231 \\  
			ARIMA  & 0.6716 & 0.5943 & 0.5389 & 0.4545 & 0.5648\\ 
			FNN-P12  & 0.6729 & 0.596 & 0.5292 & 0.4689  & 0.5667\\
			SAEs& 0.6782 & 0.6157 & 0.5553 & 0.4919 & 0.5852 \\
			DeepTransport-R1P12 & 0.6787 & 0.6114 & 0.5494 & 0.4925  & 0.5841 \\ 
			DeepTransport-R5p12 & \textbf{0.6889}& \textbf{0.6267} & \textbf{0.5724} & \textbf{0.5259} & \textbf{0.6035}\\
			\hline
		\end{tabular}
	}
	\caption{Models performance comparison at various future time points.}
	\label{tab::comp}
\end{table}

\begin{figure}[!ht]
	\begin{minipage}[b]{.5\linewidth}
		\centering
		\subfigure[Downstream attention weights]{
			\includegraphics[width=4.5cm]{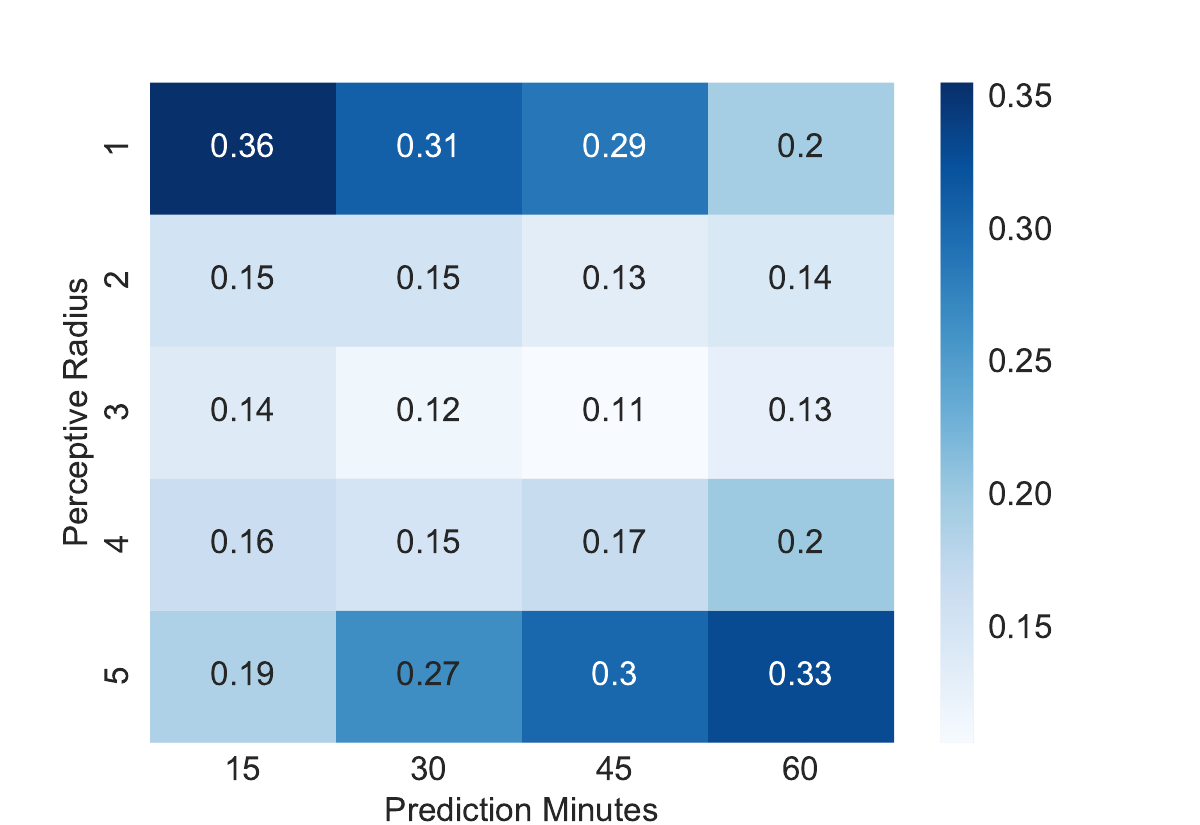}
			\label{att_down}
		}
	\end{minipage}%
	\begin{minipage}[b]{.5\linewidth}
		\centering
		\subfigure[Upstream attention weights] {
			\includegraphics[width=4.5cm]{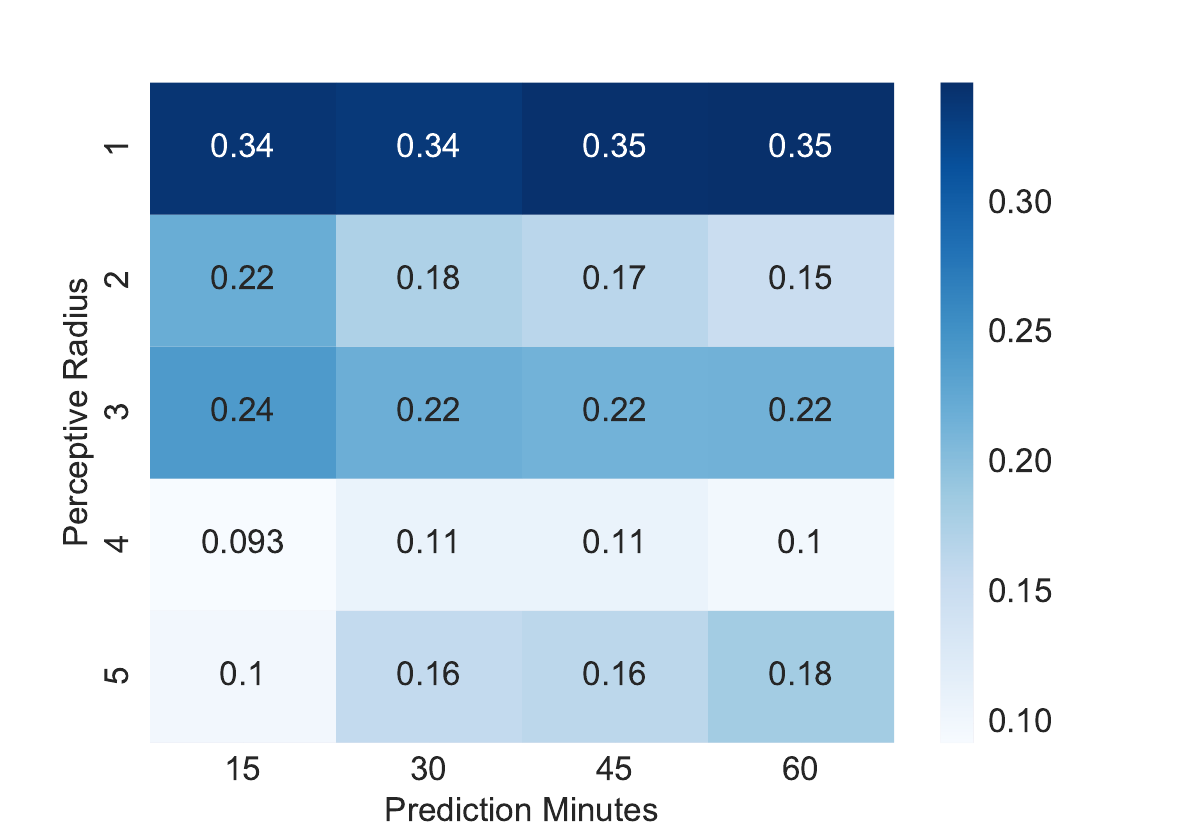}
			\label{att_up}
		}
	\end{minipage}
	\caption{Average attention weights alignments. It quantifies the spatial-temporal dependency relationships. The left figure is downstream alignments; it captures our intuition that as predicted time increased, the attention weights shifts from low-order slots to higher ones. The right figure is upstream alignments; the model pays more attention to lower orders because traffic flow in higher order is dispersed.}
	\label{fig::attention}
\end{figure}
\subsection{Slot Attention Weights}
DeepTransport also can observe the influence of each slot on the target location by checking slot attention weights. Figure~\ref{fig::attention} illustrates the attention weights between prediction minutes and perceptive radius by averaging all target locations. For downstream order slots, as shown in figure~\ref{att_down}, it can be seen that as predicted time increased, the attention weights shifts from low-order slots to higher ones. On the other side, figure~\ref{att_up} shows that the upstream first order slot has more impact on the target location for any future time. To capture this intuition, we utilized sandglass as a metaphor to depict the spatial-temporal dependencies of traffic flow. The flowing sand passes through the aperture of a sandglass just like traffic flow through the target location. For the downstream part, the sand is first to sink to the bottom, after a period, this accumulated sand will affect the aperture just like the cumulative congestion from the higher order to the lower order. Thus, when we predict the long-period condition of the target location, our model is more willing to refer to higher-order current conditions. On the other hand, the upstream part is a little different. Higher order slots are no longer important references because traffic flow in higher order is dispersed. The target location may not be the only channel of upstream traffic flow. The nearest locations that can directly affect the target location just like the sand gathering to the aperture of the sandglass. So the future condition of the target location put more attention on the lower order. Although the higher order row receives less attention in the upstream module, there is still a gradual change as prediction minutes increase.

\subsection{Case Study}
\begin{figure}[t]
	\begin{minipage}[a]{1.1\linewidth}
		\subfigure[15-minute prediction ] {
			\includegraphics[width=80mm,height=40mm]{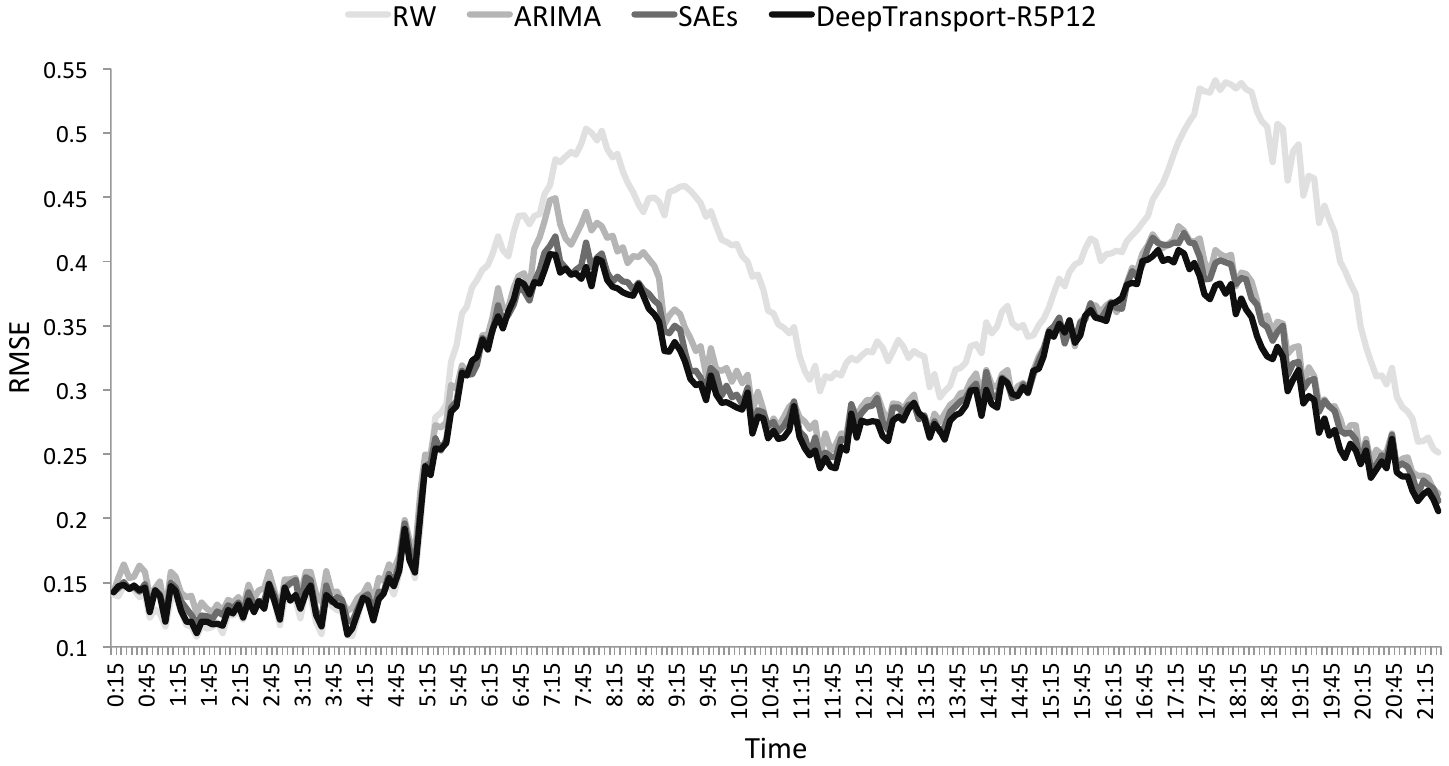}}
		\label{fig::t15}
	\end{minipage}
	\begin{minipage}[b]{1.1\linewidth}
		\subfigure[60-minute prediction ] {
			\includegraphics[width=80mm, height=40mm]{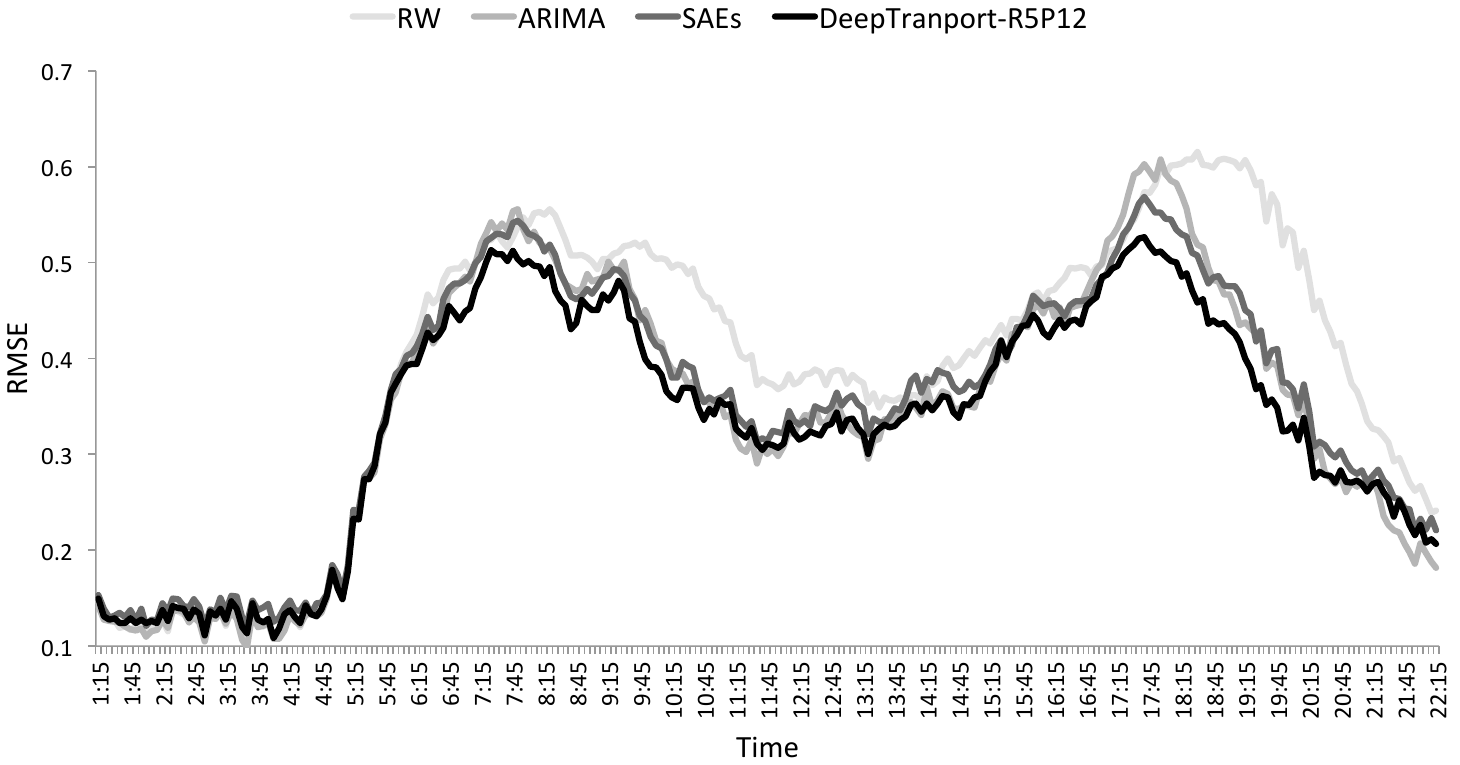}} 
		\label{fig::t60}
	\end{minipage}
	\caption{Model comparison with RMSE over time when prediction horizon equals 3 (15-minute) and 12 (60-minute)}
	\label{fig::case}
\end{figure}

For office workers, it might be more valuable to tell when traffic congestion comes and when the traffic condition will ease. We analyze the model performance over time in figure~\ref{fig::case}, which shows the Root Mean Square Error(RMSE) between ground truth and prediction result of RW, ARIMA, SAEs, and DeepTransport-R5P12. It has two peak periods, during morning and evening rush hours. We summed up three points from this figure:

\begin{enumerate} 	
\item During flat periods, especially in the early morning, there is almost no difference between models as almost all roads are fluency

\item Rush hours are usually used to test the effectiveness of models. When the prediction horizon is 15 minutes, DeepTransport has lower errors than other models, and the advantage of DeepTransport is more obvious when predicting the far point of time(60-minute prediction).

\item  After the traffic peak, it is helpful to tell when the traffic condition can be mitigated. The result just after traffic peaks shows that DeepTransport predicts better over these periods.

\end{enumerate}

\section{Related Works}
There has been a long thread of statistical models based on solid mathematical foundations for traffic prediction. Such as ARIMA~\cite{ahmed1979analysis} and its large variety~\cite{Kamarianakis2003Forecasting,Kamarianakis2005Space,Kam:space} played a central role due to effectiveness and interpretability. However, the statistical methods rely on a set of constraining assumptions that may fail when dealing when complex and highly nonlinear data. ~\cite{karlaftis2011statistical} compare the difference and similarities between statistical methods versus neural networks in transportation research.

To our knowledge, the first deep learning approach to traffic prediction was published by~\cite{huang2014deep}, they used a hierarchical structure with a Deep Belief Network (DBN) in the bottom and a (multi-task) regression layer on the top. Afterward, ~\cite{lv2015traffic} used the deep stacked autoencoders(SAEs) model for traffic prediction.  A comparison~\cite{tan2016comparison} between SAEs and DNB for traffic flow prediction was investigated. More recently, \cite{Polson2017Deep} concatenated all observations to a large vector as inputs and send them to Feed-forward Neural Networks(FNN) that predicted future traffic conditions at each location.

On other spatial-temporal tasks, several recent deep-learning works attempt to capture both time and space information. DeepST~\cite{zhang2016dnn} uses convolutional neural networks to predict citywide crowd flows. Meanwhile, ST-ResNet~\cite{zhang2016deep} uses the framework of the residual neural networks to forecast the surrounding crowds in each region through a city. These works partition a city into an $I \times J$ grid map based on the longitude and latitude~\cite{Lint2002Freeway} where a grid denotes a region. However, MapBJ provides the traffic networks in the form of traffic sections instead of longitude and latitude, and the road partition method should be considered the speed limit level rather than equally cut by road length. 
Due to the differences in data granularity, we do not follow these methods of traffic forecasting.

\section{Conclusions}
In this paper, we demonstrate the importance of using road temporal and spatial information in traffic condition forecasting. We proposed a novel deep learning model (DeepTransport) to learn the spatial-temporal dependency. The model not only adopts two sequential models(CNN and RNN) to capture the spatial-temporal information but also takes attention mechanism to quantify the spatial-temporal dependency relationships. We further released a real-world large traffic condition dataset including millions of recordings. Our experiment shows that DeepTransport significantly outperformed other previous statistical and deep learning methods for traffic forecasting.

\bibliographystyle{aaai}
\bibliography{deepTransport}



\end{document}